\documentclass[preprint]{elsarticle}
\usepackage[utf8]{inputenc}
\usepackage{graphicx}
\usepackage{subcaption}
\usepackage{booktabs}
\usepackage{amssymb}
\usepackage{float}
\usepackage{changepage}
\usepackage{multirow}
\usepackage{amsmath}
\usepackage{diagbox}
\usepackage{natbib}
\usepackage{booktabs}
\usepackage{adjustbox}
\usepackage{tabularx}
\usepackage[toc,page]{appendix}
\usepackage[inkscapeformat=png]{svg}
\usepackage{lscape} 
\usepackage{amsmath}
\usepackage{xcolor}

\title{Graph-Enhanced Model-Free Reinforcement Learning Agents for Efficient Power Grid Topological Control}

\author[add1,add2]{Eloy Anguiano-Batanero\corref{cora}}
\ead{eloy.anguiano@iic.uam.es}

\author[add2]{Ángela Fernández\fnref{fn}}
\ead{a.fernandez@uam.es}
    
\author[add1]{Álvaro Barbero\fnref{fn}}
\ead{alvaro.barbero@iic.uam.es}

\affiliation[add1]{
    organization={Instituto de Ingeniería del Conocimiento},
    addressline={C/ Francisco Tomás y Valiente 11},
    postcode={28049},
    city={Madrid},
    country={Spain}
}
\affiliation[add2]{
    organization={Departamento de Ingeniería Informática, Universidad Autónoma de Madrid},
    addressline={C/ Francisco Tomás y Valiente 11},
    postcode={28049},
    city={Madrid},
    country={Spain}
}

\cortext[cora]{Corresponding author}\ead{eloy.anguiano@iic.uam.es}
\fntext[fn]{Equal senior contributions.}

\begin{document}

\begin{abstract}
The increasing complexity of power grid management, driven by the emergence of prosumers and the demand for cleaner energy solutions, has needed innovative approaches to ensure stability and efficiency. This paper presents a novel approach within the model-free framework of reinforcement learning, aimed at optimizing power network operations without prior expert knowledge. We introduce a masked topological action space, enabling agents to explore diverse strategies for cost reduction while maintaining reliable service using the state logic as a guide for choosing proper actions. Through extensive experimentation across 20 different scenarios in a simulated 5-substation environment, we demonstrate that our approach achieves a consistent reduction in power losses, while ensuring grid stability against potential blackouts. The results underscore the effectiveness of combining dynamic observation formalization with opponent-based training, showing a viable way for autonomous management solutions in modern energy systems or even for building a foundational model for this field.

\end{abstract}


\maketitle

\section{Introduction}

Controlling a power grid is not a simple task and it is a critical aspect of providing citizens of a region with a stable and reliable power service. Each region delegates this task to a specific and highly specialized department in charge of operating the system. The main task of this department is to ensure that the grid is able to transmit the right amount of electricity from the generators or batteries to the consumers, minimizing the risk of blackout and trying to favor the use of clean and cheap energies over less suitable sources for the environment or the consumers' economy. In recent years, the increasing presence of smart grids and individual prosumers, who are both producers and consumers of energy, has added complexity to this task, often leading to unused clean energy and a growing occurrence of power curtailments.

In 2019, the french transmission system operator (TSO), named Réseau de Transport d'Électricité (RTE), began promoting the Learn To Run a Power Network (L2RPN)~\citep{l2rpn} competition aimed at advancing research into the management of these systems through the artificial intelligence field of reinforcement learning (RL). The emergence of this common framework has made this particular field quite prolific in recent years, leading the competition to continue adding new elements, such as battery management and alarm raising under specific complications, to better align with the operations of a real TSO.

The main challenge of this type of systems lies in the large number of existing terminal states. This is due to the wide range of restrictions and the vast combinatorial action space in these systems, which complicates tasks such as the effective exploration of new ways of improvement. There exist some solutions that effectively addressed the problem by providing prior expert knowledge and agent action restrictions under network security or insecurity criteria, reducing in this way the action space and the possibility that such explorations could end up in a terminal state when the network risk was not considered to be high.

In this article, we propose a new methodology for the development of this type of reinforcement learning agents that eliminates the need for expert knowledge. These strategies allow the agent to operate the network in a more autonomous way and to explore different options for reducing costs even in the nominal regime where the network is not considered to be at risk.

This paper is organized as follows. In Section~\ref{sec:soa}, we briefly describe the state of the art in using reinforcement learning to control a power network. We present the proposed method in Section~\ref{sec:proposal}, and the experimental results from 20 different situations in a 5-substation environment are shown in Section~\ref{sec:exps}. The paper ends with some conclusions and further work in Section~\ref{sec:concl}.

\section{Background and related work}\label{sec:soa}

\subsection{Power grid management}\label{subsec:powergrids}

Managing a power grid is a very complex task. Operators have to continuously monitor the elements that configure the grid to avoid undesirable situations such as blackouts while meeting the electricity demand needs of the population covered by the network. The most basic elements of a power grid are generators, loads, lines, and substations (of course, there are shunts, storage systems and other elements that complicate this task even more, but we will focus on defining just the key elements involved in this article). 

First, generators are those elements that provide us with energy. This energy can come from any type of source (wind, photovoltaic, combined cycle, etc.). They are characterized by supplying the energy that must flow through the rest of the network. 

On the other hand, loads represent the consumption points in the grid, i.e., where energy is used to satisfy the needs of the end users. These loads can vary widely in magnitude and nature, from small homes to large industrial complexes. It is essential that operators maintain a dynamic balance between the generated power and the consumed one, as any imbalance could lead to voltage or frequency fluctuations that compromise the stability of the grid.

Transmission or distribution lines allow power to flow between generators and loads. These lines are designed to transport large amounts of electricity over long distances, minimizing energy losses in the process. However, factors such as material resistance, weather conditions and overloads can affect their performance or even induce blackout situations when unwanted connections are made (for example, a cable getting too hot can cause it to bend and may potentially contact with vegetation under the line), requiring constant maintenance and meticulous planning.

Finally, substations play a crucial role in the transformation and distribution of energy. These facilities allow voltage levels to be changed so that electricity can be transported efficiently, as well as a series of bus bars that allow the electrical circuit to be configured at each instant of time in the network. In this paper we will focus on the management of an electrical network with substations featuring only two bus bars, which means that the connection of each of the elements mentioned above can be to bus number 1, bus number 2, or disconnected by isolating or grounding it.

A fairly common way of operating power grids is managing topology by controlling the connections of elements to different substation buses to avoid undesirable situations, but it is not the only one. In addition, the system operator usually has generation and load forecasts at the different points of the grid, allowing it to communicate with the managers of those generators to adjust generation values in a prescriptive way. This is when two new types of management appear: curtailment and redispatching.

Curtailment refers to the intentional limitation of the generation of certain resources, usually renewable such as wind or solar energy, when the grid cannot accommodate all the energy generated due to transmission capacity constraints or a temporary imbalance between generation and demand. This type of management is usually a preventive measure to avoid overloads in lines or substations, and although it results in a loss of energy that could have been used, it is preferable in the face of a possible instability or failure in the system. Conversely, redispatching involves the opposite action of a curtailment. In this case, operators can request specific generators to increase or decrease their output to a specific value, depending on the location of the problem on the grid and the available resources elsewhere in the system. Although more costly, this process optimizes energy flow and avoids critical congestion, ensuring operational stability. It is important to note that not all generators are susceptible to redispatching or curtailment. For instance, it is possible to request more power from a combined cycle power plant, but energy sources based on natural resources, such as wind or solar, cannot be controlled in this way. Similarly, there are generation elements for which, although curtailment is technically possible, it is considered virtually impossible due to the technical complications involved, such as in the case of a nuclear power plant.

\subsection{Reinforcement learning: an introduction} \label{subsec:rl}
In 1938, B.F. Skinner coined the psychological term “operant conditioning”~\citep{skinner1938behavior} as a method of learning as opposed to Pablov's classical conditioning. The term refers to a form of learning based on an agent interacting with its environment and being guided toward a particular goal through a system of reward or punishment, without the need for external knowledge. Years later, in the field of computer science, this mode of learning was abstracted as an iterative trial-and-error process in which the environment has to be formalized as a Markov Decision Process (MDP) or a Partially Observable Markov Decision Process (POMDP).

An MDP is a discrete stochastic control process aimed at optimizing decision-making, where a set of states $S$, a set $A$ of actions to take, a reward function $R(a_t ,s_t, s_{t+1})$, and a state transition function $F(s_{t+1}|s_t, a_t)$ are defined. Thus, a generalization of the function to optimize in a process of this type would be

\begin{equation*}
    \mathbb{E}\left[\sum_{t=0}^{\infty} \gamma^t R(a_t ,s_t, s_{t+1}) \right],
\end{equation*}

where $a_t$ is the action chosen by $a_t = \pi(s_t)$, $\pi$ is the optimization policy that we are trying to look for and $\gamma \in (0, 1)$ is used as a discount factor for future rewards and controlling long-time planning of the agent.

Within the multitude of agent types that can be trained, there is a specific family called Actor-Critic agents. This family of algorithms simultaneously trains two flows with parameters $\theta$ and $\phi$, in such a way that the policy $\pi_{\theta}$ will be the actor and therefore the one in charge of choosing the action at a given time, and on the other hand, the value function $V_{\phi}$ is estimated, which tries to determine the amount of reward that remains to be obtained by the agent until the end of the trajectory.

\subsubsection{PPO algorithm with mask over actions.}\label{subsubsec:ppo}
One of the most widely used methods of the Actor-Critic family in the literature is Proximal Policy Optimization (PPO)~\citep{Schulman2017ProximalAlgorithms} due to its good results compared to other types of strategies. Firstly, it utilizes the concept of advantages, represented as $A_{\phi}(s_t, a_t, \pi) = \delta_t + \gamma \delta_{t+1} + \gamma^{2} \delta_{t+2} +... + \gamma^{T-2} \delta_{t+T-2}$ where $\delta_{t} = r_t + \gamma V_{\phi}(s_{t+1})-V_{\phi}(s_t)$. Advantages assess whether certain actions yield higher or lower rewards than expected, guiding their encouragement or discouragement. This concept is further generalized through Generalized Advantage Estimation (GAE), defined as $A^{GAE(\gamma, \lambda)}_{t} = \sum\limits^{\infty}_{l=0} (\lambda \gamma)^l \delta_{t+l}$
Here, $\lambda$ balances the trade-off between bias and variance as it is used as an exponential discount weight.

Additionally, PPO introduces an intrinsic reward mechanism via an entropy term to promote exploration during the agent's training. Furthermore, the policy loss in PPO incorporates a clipping mechanism that acts as a regularizer. This feature prevents substantial policy changes between iterations, ensuring smoother and more stable policy updates. The original PPO algorithm has this loss function definition
\begin{equation*} 
   L(\theta)=L^{clip}(\theta) -c_1 L^{VF}(\phi)+ c_2 S(\theta),
\end{equation*}
with the following entropy bonus loss to encourage exploration
\begin{equation*}
   S(\theta) = \mathbb{E}_t\left[-\pi_\theta (a_t, s_t) \log(\pi_\theta(a_t, s_t))\right],
\end{equation*}
the value function (or critic) loss defined as
\begin{equation*}
    L^{VF}(\phi) =  \mathbb{E}_t \left \| r_t + \gamma V_{\widehat{\phi}}(s_{t+1}, \pi)) -V_{\phi}(s_{t}, \pi)) \right \|^{2}_2,
\end{equation*}
with an old copy of the value function estimator with parameters $\widehat{\phi}$, and the policy (or actor) loss as
\begin{equation*}
    \begin{gathered}
        t_1 = A_{\phi}(s_t, a_t, \pi) R_t(\theta), \\
        t_2 = A_{\phi}(s_t, a_t, \pi) clip(R_t(\theta), 1-\epsilon, 1+\epsilon),\\
        L^{clip}(\theta) =  \mathbb{E}_t[\min(t_1, t_2)],
    \end{gathered}
\end{equation*}
where $R_t = \pi_{\theta}(a_t, s_t)/\pi_{\widehat{\theta}}(a_t, s_t)$, and $\pi_{\widehat{\theta}}$ is the policy from the preceding iteration. This clipping mechanism mitigates drastic policy updates, effectively regularizing the optimization process.

One of the most interesting features of this algorithm for our case is that a system for masking invalid actions can be easily implemented according to the current observation of the environment. This mechanism essentially assigns a null value to the probability of sampling the invalid action at the output of the logits of each of the possible actions~\citep{MASKABLEPPO}. The masking mechanism is defined as 

\begin{equation*}
    P[A|L(s_t)] = softmax(L(s_t))
\end{equation*}

where the masked logits values are calculated as

\begin{equation*}
    L_{i}(s_t) = \begin{cases}
        \pi_{\theta}(A|s_t)_i & \quad M(s_t)_i = 0 \\
        -\infty  & \quad M(s_t)_i = 1\\
       \end{cases}
\end{equation*}

and $A$ is the set of actions, $\pi_{\theta}(A|s_t)$ is the policy, $s_t$ is the state at time $t$, and $M(s_t)$ is the vector mask at time $t$. 

This is especially useful in our case, since as will be specified later in section \ref{subsec:grid2op}, an environment intended to simulate an electrical network has multitude of invalid actions that can be avoided thanks to this mechanism.

\subsection{Grid2Op framework} \label{subsec:grid2op}
Grid2Op~\citep{grid2op} is an open-source framework proposed by RTE and utilized in the L2RPN competition to abstract researchers from the complexity of the state transition function $F$ that simulates a powergrid, following some of the standards proposed by OpenAI's Gym~\citep{gym} or Gymnasium~\citep{gymnasium} libraries. In the field of reinforcement learning, it is common to refer to this abstraction of the transition function as the ``environment''. Additionally, this framework provides programmers and researchers with the freedom to establish the definition of a formalization of a state of the power grid $s_t$ (also called ``observation''), as well as the set of actions $A$ that the agent can take and the reward function $R$ between states of the grid.

The Grid2Op framework makes several assumptions about the MDP definition over power networks that must be taken into consideration, as they are key to both the literature and this article. The first of them concerns the conditions under which a terminal state has been reached, i.e. the episode has ended:
\begin{itemize}

    \item Grid2Op operates by testing power networks through situations based on historical datasets of generation and load. These situations are referred to as ``chronics'' and have a maximum duration of 2016 steps (in the case or the \textit{rte case5 example} environment) of 5-minute intervals. Additionally, when configuring each chronic, a maximum number of 5-minute steps that the agent can perform on that chronic can be specified. If either of these two values is reached (end of chronic or maximum iterations), the episode is finished, making the desired terminal state to be reached in any case.
    
    \item Power networks have the peculiarity of being highly unstable. A total network collapse is referred to as a ``blackout''. A common criterion for evaluating the stability of a power network is the ``N-1 criterion''. This criterion is defined as the resilience of the network after taking an action, i.e. when removing any element from the network's topology the result should still be equally stable. If the agent leads either to a blackout state or violates the N-1 criterion, the environment provided by Grid2Op will end that episode.

    \item As these systems aim to provide the power from generators to loads, whenever a load is not satisfied, the environment treats this as an end of an episode as this means that the main objective of the power grid is not met.
    
\end{itemize}

\subsubsection{Chronics and opponents}
Simulating the activity of an energy network is a complex process, as it depends on specific load and generation data at various points in the grid and the topology of the grid itself. To address this, the Grid2Op framework introduces the aforementioned concept of ``chronics''. Chronics are predefined datasets containing generation and load scenarios that unfold step by step during a simulation. These scenarios allow for the reproducibility of specific conditions, making them a valuable tool for testing and evaluating the performance of agents under consistent and diverse conditions, without relying on random or unrealistic events.

A collateral effect of having an environment based on static data is the little variation to which the agent can be subjected as a result of always facing the same situations. To mitigate this effect, the framework also provides us with a very important tool:  opponents. Opponents are configurable tools that allow us to have a process that makes the agent's task more difficult, for example, by making a powerline unavailable without prior warning by ``attacking'' it. This strategy helps to mitigate a possible lack of generalization of situations in the data and helps trained agents to be more robust to adverse or unforeseen situations.

\subsubsection{Actions}
Another challenge with such systems is the large number of possible actions and the complexity in sampling one action coherent with the current state of the network. To address this, the framework provides information on whether the action has been categorized into four distinct groups: correct, illegal, ambiguous, or erroneous. 
A correct action is one that the Grid2Op framework can easily interpret and execute without issues. This is the type of action our agent should produce, as it represents a well-structured and coherent decision. On the contrary, an erroneous action would be one that leads the environment, simulator, or transition function of the environment to an unsolvable state for the power flow solver used by Grid2Op as backend. While it is interesting to avoid such actions, they may still occur due to internal mathematical complications inherent in any power network load flow simulator. The remaining groups of actions (illegal and ambiguous) present more subtle differences. According to \citet{ActionsGrid2op}, illegal actions are those that are not feasible given the current circumstances of the power network (e.g., reconnecting a powerline that is under maintenance), while ambiguous actions are those that do not make sense by themselves (e.g., connecting element 999 of the network, which does not exist). In this sense, we must ensure that ambiguous actions do not arise from a well-designed action space, while illegal actions should either be avoided by encoding the logic to prevent the agent from taking such actions or by discouraging the agent from doing so through the design of the reward function.

Correctly understanding the differences between these action categories under this framework is a key insight for one of the major challenges agents face in properly managing a power network, as well as for understanding much of the developments and proposals that will be discussed later in this article.

\subsection{Previous work}

This problem is not new, as there have been different L2RPN competitions over the last few years, so there is literature on the subject in which different approaches have been proposed to address each of the competitions and its particularities. It is important to note that the competition tends to change its bases and terms frequently, so some solutions will focus more on solving certain problems than others. 

In \citet{9128159}, an expert system capable of discovering actions in critical states using a heuristic of paths over graphs is formalized, thus using expert knowledge to establish a methodology. After the generation of this first expert system, the problem was addressed as a reinforcement learning problem in \citet{lan2019aibasedautonomouslineflow}. In this approach, imitation learning is used to guide an exploration of Dueling DQN agents \citep{DBLP:journals/corr/WangFL15} with an early warning mechanism to favor ``Do Nothing'' actions. Subsequently, \citet{10.1007/978-3-030-86517-7_11} propose a Q-learning search-based planning approach , enabling an agent to perform the top-k chosen actions and achieve the victory in the robustness track of the L2RPN competition. Similarly, \citet{MAROT2022108487} introduce the concept of alarms, and by incorporating Charles Green’s Trust Equation, they allow the agent to identify and warn about possible problematic areas. Moving on to more sophisticated learning techniques, in \citet{Dorfer2022PowerGC} they use the adaptation of AlphaZero~\citep{silver2017masteringchessshogiselfplay} to a single player infinite horizon game called MuZero~\citep{Schrittwieser2020}. They propose a two-hierarchical decision model to achieve target topologies using afterstates as input to the model. Recently, in \citet{MARL} they set a series of actions on each of the substations in a network using expert knowledge and the Corrective and Preventive Actions (CAPA) policy to address the problem as a multi-agent system.

\section{Our approach} \label{sec:proposal}

Our approach is an attempt to make some simplifications of the problem, aligning it with a classical reinforcement learning problem, and to investigate the possible limitations of various assumptions that have been made in the past. We seek to reformulate the problem in such a way that the dependence on expert knowledge is minimized and to be more efficient avoiding, for example, the simulation of distinct future states of the grid present in prior works, due to the extra costs that this implies, as it involves a series of power flow solutions for the different network states.

In order to achieve this, we propose the definition of a model-free reinforcement learning problem without falling into the need for future simulations or afterstates (states for the already applied action also called post-decision states) due to the extra computational cost this carries. To this end, some of the key points will be an action space with correct sampling according to the current state of the network, correct state formalizations, and dynamic masking of undesirable actions, as it will be explained in the following subsections.

\subsection{Observations}
As we have discussed previously, there is no single way to formalize each observation (the state $s_t$) of a power grid, but it is crucial for the good performance of an agent, as it will serve as the input of the model. Specifically, the assumption of one formalization over another can enable the agent to better distinguish one situation from another, and thus yield better results, as it serves as the entry point to the decisions the agent can or cannot make. In this study, various alternatives will be presented for comparison, in order to discuss their advantages and disadvantages.

\subsubsection{Flat Observation} \label{subsubsec:flat}
The formalization of the observation called ``Flat'' is the simplest one. In this case, a feature vector is constructed to summarize the entire current state of the network. These features include:

\begin{itemize}
    \item Voltage modules of each energy producer (generators and batteries).
    \item Power modules, active power components, and reactive power components of each producer and load in the powergrid.
    \item Capacity of each power line (rho), representing the current flow relative to the thermal limit of the line.
    \item State of each line (connected/disconnected).
\end{itemize}

\subsubsection{Substation Graph}
\begin{figure}[H]
    \centering
        \includegraphics[width=\textwidth]{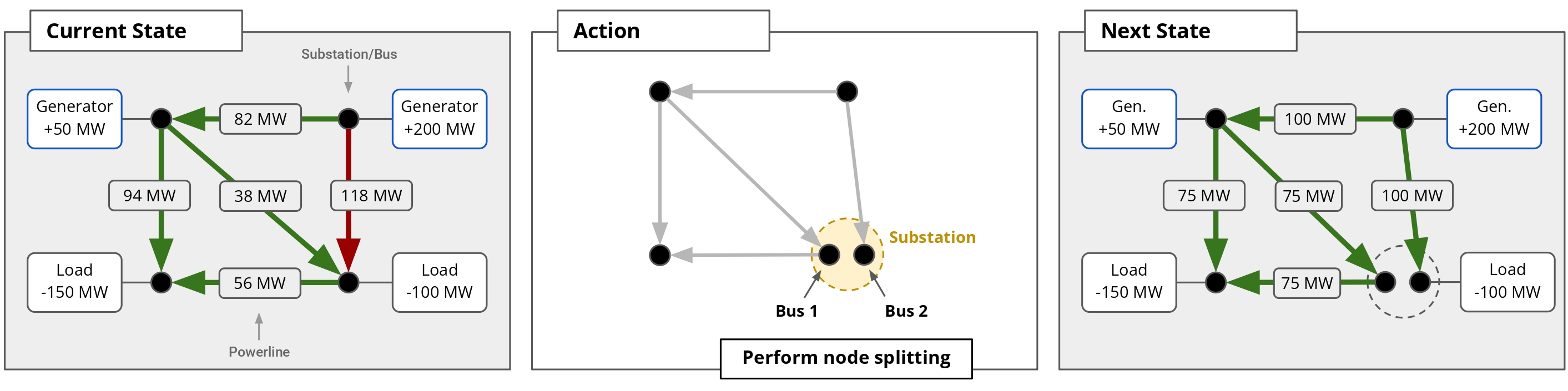}
    \caption{Illustrative example on how a powergrid can be formalized as a dynamic graph with a variable number of nodes, since each unique connection of each substation would be seen as a node of the network. Thus, when performing certain actions on the network, it is possible for a substation to correspond to more than one node as a previously inactive bus would come into play. This figure is extracted from~\citet{Dorfer2022PowerGC}.}
    \label{fig:substation_graph}
\end{figure}

The substation graph presents a more advanced way of formalizing the states of the power grid. An electrical substation is an installation in which the different connections to the busbars are specified, thus configuring the topology of the substation and, ultimately, of a power network, is critical. In the particular case of the network that will be addressed in this paper, we have substations with two busbars, which indicates that the elements associated with each substation can be connected to bus 1, bus 2 or ground, which would be the equivalent of disconnecting the element from the network. Therefore, the substation graph can be seen as a dynamic undirected graph where each observation will correspond to a variable number of nodes, and the connections between them will not have a specific direction. In this context, a node is assumed to represent the unique connections occurring at a substation and an edge would be one or more powerlines between those unique connections. A more graphical explanation can be seen in Figure \ref{fig:substation_graph}.

It is important to note that this formalization, besides providing a good topological approximation of the power grid state, also allows us to define a set of features for each of the resulting nodes, as well as features associated with the edges, as shown in Table~\ref{tab:substation_graph_variables}.

\begin{table}[H]
    \small
    \centering
     \caption{Features corresponding to the nodes and edges of the dynamic substation graph.}
    \label{tab:substation_graph_variables}
    \begin{tabular}{l|l|l}
        Feature & Origin & Description \\
        \midrule
        \midrule
        $|p|$ & \multirow{6}{*}{Nodes} & Power module \\
        $p$ &   & Active power component \\
        $q$ &   & Reactive power component \\
        $|v|$ &   & Voltage module \\
        $sin(\theta)$ &   & Sine of voltage angle \\
        $cos(\theta)$ &   & Cosine of voltage angle \\
        \midrule
        $|p|$ & \multirow{10}{*}{Edges} & Power module \\
        $p$ &   & Active power component \\
        $q$ &   & Reactive power component \\
        $|v|$ &   & Voltage module \\
        $sin(\theta)$ &   & Sine of voltage angle \\
        $cos(\theta)$ &   & Cosine of voltage angle \\
        $\rho$ &   & Capacity of the powerline \\
        $p_{tsoverflow}$ &   & Timestep overflow \\
        $p_{tscooldown}$ &   & Time before cooldown powerline \\
        $p_{maintenance}$ &   & Duration of next maintenance \\
        $n_{lines}$&   & Number of connected power lines represented by this edge \\

    \end{tabular}
\end{table}

\subsubsection{Element Graph}\label{subsubsec:element}

Continuing with the approach of viewing the power grid as a graph and based on previous state-of-the-art approaches to power grid optimal power flow solvers~\citep{LOPEZGARCIA2023105567}, we propose the use of the Element Graph. The Element Graph is a graph in which each element of the network (generator, load, line, bus, storage, or shunt) is a node, and the edges only specify whether a node could be connected with another (in this case, an edge feature just specifies if the state is connected or disconnected). Thus, a single power grid will be an undirected graph of fixed size.

\begin{figure}[H]
    \centering
        \includegraphics[width=1.1\textwidth]{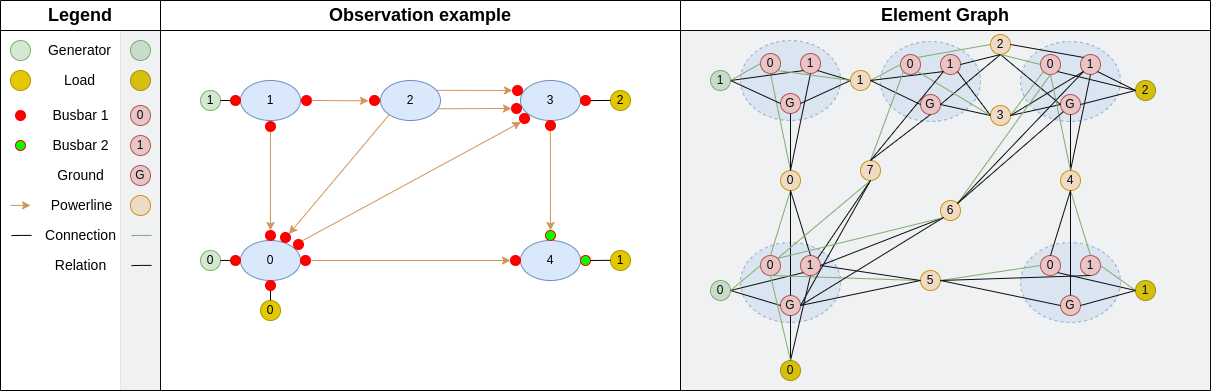}
    \caption{Illustrative comparison of an observation without formalization (a powergrid state) of 5 substations against its equivalent in Element Graph form. It can be seen how each element always exists in the graph, as well as how all elements share an edge with each element to which it could be connected.}
    \label{fig:element_graph}
\end{figure}

Figure \ref{fig:element_graph} shows how a power network of 5 substations (the simplest, but also the one that will be used in the experimental phase at Section \ref{sec:exps}) is transformed from a human-understandable scheme without any mathematical formalization to an Element Graph formalization understandable by deep reinforcement learning agents. The resulting node features for this graph are shown in Table~\ref{tab:element_graph_variables}.

\begin{table}[H]
    \small
    \centering
    \caption{Node features used for each type of node in the case of the formalization called ElementGraph. It can be seen that there are some features that appear in all elements, but others are specific for each node type. }
    \label{tab:element_graph_variables}
    \begin{tabular}{l|l|l}
        Variable & Node Type & Description \\
        \midrule
        \midrule
        $|p|$ & \multirow{5}{*}{All} & Power module \\
        $p$ &   & Active power component \\
        $q$ &   & Reactive power component \\
        $|v|$ &   & Voltage module \\
        $cos(\theta)$ &   & Cosine of voltage angle \\
        \midrule
        $g_{norm}$ & \multirow{8}{*}{Generator} & Normalized generation power ($(|p|-|p_{min}|) / (|p_{max}| - |p_{min}|)$)\\
        $g_{max up}$ &   & Generator max ramp up \\
        $g_{max down}$ &   & Generator max ramp down \\
        $g_{min uptime}$ &   & Generator min uptime \\
        $g_{min downtime}$ &   & Generator min downtime \\
        $g_{cost}$ &   & Generator cost per MW \\
        $g_{start cost}$ &   & Generator startup cost \\
        $g_{shutdown cost}$ &   & Generator shutdown cost \\
        $g_{type}$ &   & One-hot generator type (solar, wind, hydro, thermal or nuclear) \\
        \midrule
        $b_{type}$ & \multirow{2}{*}{Bus} & One-hot busbar (ground, busbar1, busbar2) \\
        $b_{coooldown}$ &   & Cooldown time \\
        \midrule
        $\rho$ & \multirow{4}{*}{Powerline} & Capacity of the powerline \\
        $p_{tsoverflow}$ &   & Timestep overflow \\
        $p_{tscooldown}$ &   & Time before cooldown powerline \\
        $p_{maintenance}$ &   & Duration of next maintenance \\
    \end{tabular}
\end{table}

\subsection{Model architectures}

The agents were implemented with the reinforcement learning library Stable Baselines3~\citep{stable-baselines3} (SB3), which can not natively work with environment states in the form of a graph as it can only work with vectorized observations. Thus, we require of a method for the vectorization of a state and observation of the power network. This leads us to develop feature extractors according to each of the state formalizations we propose.

First we have the Flat Observation formalization which, as its name indicates, is already a vector, so the only thing that will be necessary is to make the transformations with a fully-connected network or a multilayer perceptron that projects the data into a pre-fixed dimensionality vector space, so SB3 can use it normally.

However, the other two formalizations (SubstationGraph and ElementGraph) require addressing the challenge of generating a latent vector that encodes the node, edge and topology variables of a particular power network state. To this end, the field of Graph Neural Networks offers different model architectures such as Graph Convolutional Networks~\citep{GCN}, Gated Graph Convnets~\citep{GatedGCN} or Graph Atention Networks~\citep{GAT}. In our particular case, we chose GraphTransformer~\citep{GraphTransformer}. This model architecture generalizes traditional transformer networks to accommodate graphs by incorporating essential features of graph structures. In particular, this architecture includes Laplacian eigenvectors for node positional encodings, which replaces the sinusoidal encodings commonly used in the Natural Language Processing (NLP) field. This allows the model to effectively capture the relative positions of nodes within the graph. 
\begin{figure}[H]
    \centering
        \includegraphics[width=0.8\textwidth]{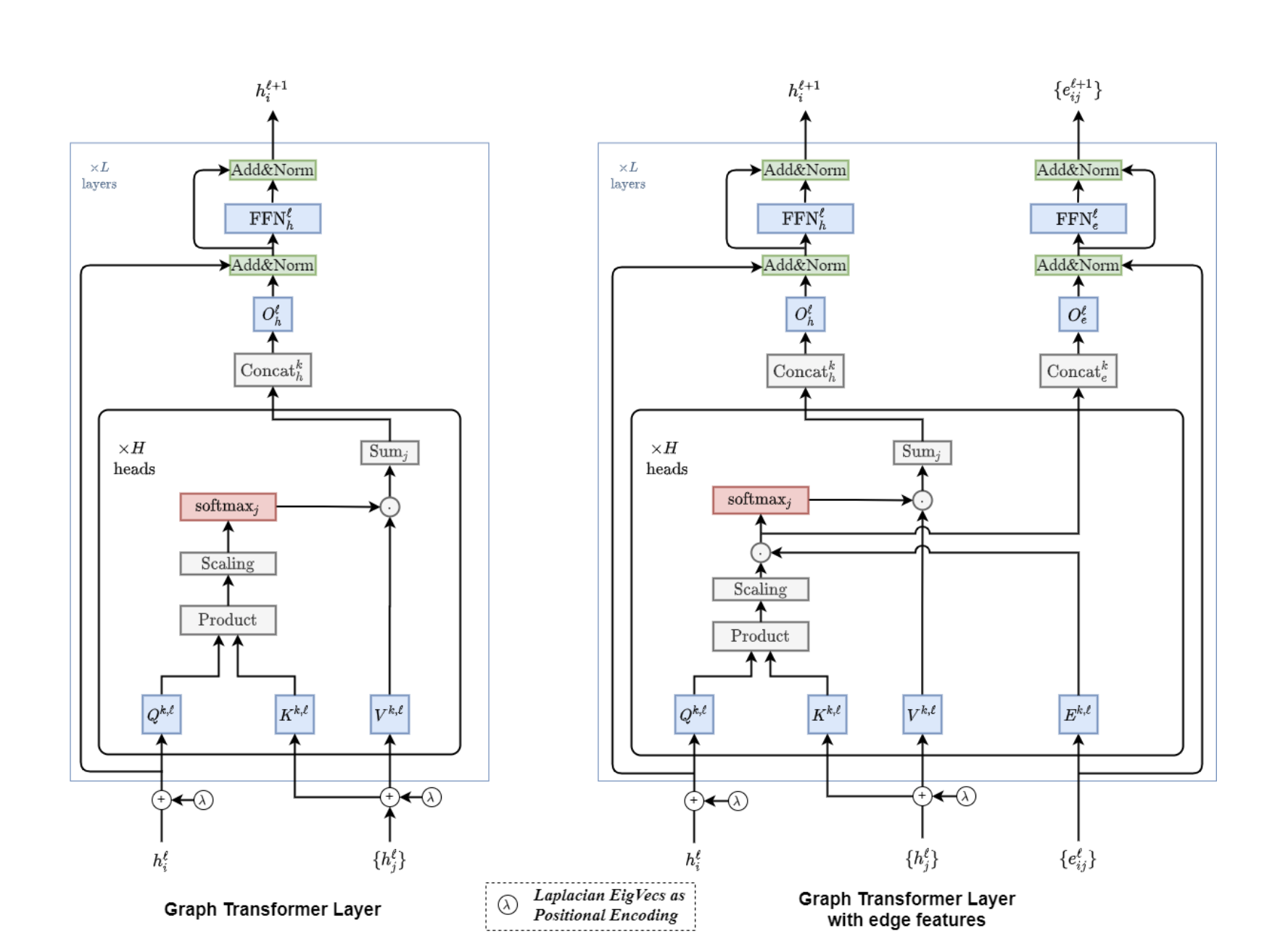}
    \caption{Graph Transformer architecture proposed by \citet{GraphTransformer}, highlighting its integration of neighborhood connectivity, Laplacian eigenvector positional encodings, and edge feature representation to enhance graph-based learning tasks.}
    \label{fig:graphtransformer}
\end{figure}

Furthermore, the architecture supports the inclusion of edge features, enabling it to handle tasks that need understanding relationships beyond just node connections. By addressing these key elements, the Graph Transformer architecture is a potent alternative to both traditional transformer models and graph neural networks (GNNs), demonstrating strong performance across various graph-based applications.

In the particular case of wanting to obtain a single latent vector that summarizes the entire graph to be able to use this architecture as a SB3 feature extractor, the original paper proposes a mean-readout operation of the individual encoder output latent vectors. Assuming a graph as $G = (V, E)$ with nodes $v_i \in V$, edges $e_{ij} \in E$ and $N$-dimensional node features $h_i \in \mathbb{R}^N$, the GraphTransformer applies its attention mechanism and projects each $N$-dimensional $h$ vector into a $D$-dimensional $h'$ vector where $D$ is the chosen hidden size as hyperparameter. Then, the readout operation takes the form
\begin{equation*}
    MeanReadOut(H')= \frac{1}{\lvert V \rvert}\sum_{i}^{V} h'_i.
\end{equation*}

Building on top of this, in this work slight modifications have been made to the GraphTransformer architecture. The positional encoding of the graph nodes is computed through a linear projection of basic graph structure features (input centrality and output centrality) instead of the Laplacian Positional Encoding proposed by the original paper. Also, in the case of the Element Graph observation described in Section \ref{subsubsec:element}, different linear projections prior to the GraphTransformer have been added, specialized according to the electrical grid node type . This decision is supported by the advantages of using a TypedGNN over this formalization of an electrical grid described in~\citep{LOPEZGARCIA2023105567}, which uses it to solve a power flow problem with a graph-based deep learning model.

\subsection{Actions}
As explained above in Section \ref{subsec:rl}, one of the main challenges of this problem is to ensure that the action to be applied at a particular step makes sense topologically and is in accordance with the current state of the network in question. For example, trying to change the state of the power line number 999 would be impossible as it does not exist and it will be specified as an illegal action by Grid2Op framework \ref{subsec:grid2op} or you could even try to change the status of an element that is in a period in which it cannot be modified (it is possible that it is under maintenance, that it still has cooldown time left because it has been changed recently, etc.). To avoid having to disincentivize this type of actions through a minor or even negative reward, it is necessary to address this problem from the design of the environment's action space,  preventing such actions and facilitating the agent's training.

Therefore, a maskable Topological Action Converter (TAC) has been developed in which each network element manages each of the possible connections to the busbars of a particular substation. For a generator assigned to a substation, this action dimension will have 4 possibilities: connection to ground or disconnection, do nothing (useful for the use of masks), connection to busbar 1 and connection to busbar 2. On the other hand, for a load we will have only 3 possibilities, since the disconnection option does not make sense since the environment assumes that not meeting an electrical demand constitutes a terminal state and does not take into account Value of Lost Load (VOLL)~\citep{GORMAN2022107187} situations. Finally, in the case of power lines there will be 6 possibilities: grounding or disconnecting, doing nothing and the 4 combinations of connecting to busbars 1 and 2 at each of their ends as power lines connect one substation to another, so each end has a different bus connection at each substation.

\begin{figure}[H]
    \centering
        \includegraphics[width=0.8\textwidth]{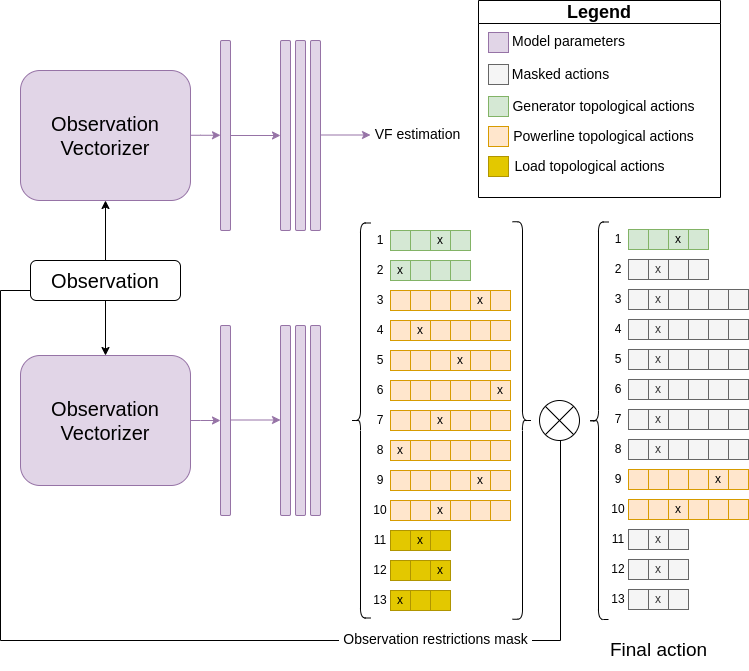}
    \caption{In this figure it is shown how the PPO algorithm with masks from Section \ref{subsubsec:ppo} works along with the TAC. This algorithm uses the observation vectorizer to optimize both the value function and the agent's policy since it is an algorithm of the Actor-Critic family. The action of the policy $\pi_{\theta}$ would be the final action on the image while the VF estimation is the output $V_{\phi}$ of the Critic that tries to estimate the incoming rewards from now on. On the other hand, it can be seen how after the decision of the agent's policy, the masking of actions allows us to avoid the appearance of illegal actions taking into account the logic and rules of the current observation by forcing the dimension assigned to do nothing on that element on masked ones.}
    \label{fig:tac_actions}
\end{figure}

Figure \ref{fig:tac_actions} shows an example of how the observation vectorizers are used in conjunction with the TAC to allow the agent to choose an independent topological action of each element, which will pass a subsequent filtering using a mask based on the rules of the actual environment state to make it impossible for an illegal action to appear. The mask is constructed in such a way that the agent can only choose the do-nothing option for each masked element whenever the element is not available as an actionable one.

\subsection{Reward}
\label{sec:reward}
The problem we are concerned with is categorized as a control problem. This means that its operation directly influences the duration of the episodes, making this a critical point to solve the problem, since one of the additional objectives is to maintain the system active or functioning. As such, when using a reinforcement learning approach, constraints have to be applied in the reward function designed for the objective in question. For example, a constraint on the design of such function would be that it should be mostly positive at each step of the MDP and not excessively penalizing in cases where it is not positive. Otherwise, the agent would run the risk of learning to force the system to crash early in the episode in order to assume a reward that, although negative, at least does not continue to become more and more negative as the episode progresses.

Therefore, we have designed a reward function that is positive in most of the situations, not excessively negative in the rest, and additionally we have managed to limit the maximum reward obtainable in each of the steps. Such reward function is defined as
\begin{equation*}
R(s_t, a_{t-1}) = 
     \begin{cases}
       0  &\quad \text {if } a_{t-1} \in [A_{Error} \cup A_{Illegal}] \\
       1  &\quad \text{if }MaxLoss_{s_t} = 0\\
       \frac{(MaxLoss_{s_t} - Loss_{s_t})}{MaxLoss_{s_t}} & \quad \text{otherwise,}\\ 
     \end{cases}
\end{equation*}
where $MaxLoss_{s_t} = M_{s_t} \sum_{g \in G} power_g$ and $Loss_{s_t} = M_{s_t} (\sum_{g \in G}power_g - \sum_{l \in L} load_l)$, $G$ are the generators, 
$L$ are the loads, and $M_{s_t}$ is the marginal cost of electricity per MWh. This formulation allows us to obtain a bounded and mostly positive reward, so the agent must reduce the amount of network losses to maximize it.
\section{Experiments} \label{sec:exps}

\subsection{Experimental setup}

As mentioned previously, this type of environment relegates its operation to a series of generation and load datasets, in which a network situation is presented and the network operation is simulated. These series of generations and loads are called ``chronics''.  Also, it has been decided to set the agents that do not perform any action (``Do Nothing'') as a baseline to beat, since the metrics used by the competition establish the value $0.0$ as the reward obtained by an agent that does nothing at each step of the chronics. Additionally, it is important to note that this comparison is not trivial since inaction is an important part of network operation if the change is not expected to convert into a clear future gain, so it is an action that is usually advantageous and quite common.

In the chosen experimental environment, we have five substations with 20 different chronics, numbered from 0 to 19 and with 2016 steps per chronic. First of all, it has been decided to reserve chronics 17 and 19 as pure testing chronics, while the rest will be used also in the training phase. This decision is based on the split made in~\citet{MARL}, where both are marked as two really complicated chronics for the ``Do Nothing'' agent.

To add variability to the scenarios that can be seen by the agent and to help isolate the effects of each of the actions on the current reward, it has been decided to limit each trajectory to 864 steps (3 days in 5-minute lapses) and to set different trajectory starting points for each of the chronics, with each one starting from 0 to 4 days after its beginning. This additional variability gives us 90 possible training environments (18 train chronics with 5 possible jumps each). These environments are sampled uniformly at training time.

In the test phase, the exploration of the agents has been disabled. This means that the agent becomes deterministic and at each step the selected action is the one with the greatest policy value. Also, each of the available chronics of the environment has been used to check the end-to-end performance throughout the training process, being aware that the agent in question has not been able to observe chronics 17 and 19. In addition, to check the robustness of the resulting agents, each of the 20 chronics has been evaluated with and without opponent (See Appendix \ref{appendix:results}).
 
\subsection{Rewards optimization}

The methodology explained above implies a multitude of different scenarios for training and evaluating our agents. Among all of them, in this experiment we have defined the scenarios in which an agent has been trained or not over environments with opponents. Also, for every agent, the evaluation metrics are obtained over every chronic (train and test ones).

\begin{figure}[H]
    \centering
        \includegraphics[width=\textwidth]{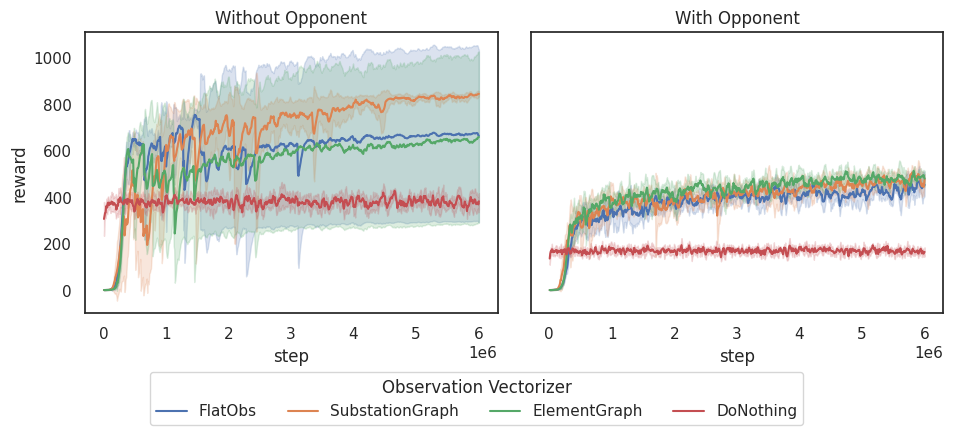}
    \caption{Rewards obtained at training chronics throughout the training of each agent configuration. Each proposed approach was run 5 times and the bands show the standard deviation of that approach. The left plot shows agents trained without an opponent, while the right plot presents results when the opponent is active during training.}
    \label{fig:train_rewards}
\end{figure}

Firstly, we analyze how well the presented agents optimized the rewards used as targets during their training (section \ref{sec:reward}).
In Figure \ref{fig:train_rewards} we can see the training curves of the different approaches considered: agents training with and without opponents and different observation converters. The results indicate that training phases including an opponent lead to lower reward values, which is expected due to the increased difficulty and variability introduced by the opponent. However, these curves also show significantly lower variance, suggesting that training with an opponent produces more stable and robust policies.

\subsection{Stability analysis}

Due to the existence of great variability in the evaluation of this type of agents (chronics, opponents, etc.), it has been decided to create a metric that takes into account the convergence and stability of the agents that are evaluated. This metric is called ``Steps to Complete'' (S2C): intuitively, it is the number of training steps required by the agent to learn a stable strategy capable of keeping the network running throughout the duration of a complete chronic without fail. More formally, S2C is defined as the number of steps that the agent has needed to complete the 2016 steps of each chronic without opponent and with the restriction of not failing to complete it in future evaluations during the life cycle of each agent. The formulation of this metric would be 
\begin{equation}
  S2C(c) = argmin_s(\tau(\theta_k, c) = |c| \, \forall k \ge s),
\end{equation} \label{eq:s2c}
where $c \in C$ is the chronic being evaluated, $C$ is the chronics set, $\tau(\theta_k, c)$ is the length of the episode at $c$ following  $\pi_{\theta_k}$,  $|c|$ is the step length of chronic $c$, $s \in S_{TRAIN}$ is the number of the train step, $S_{TRAIN}$ are the training steps and $\pi_{\theta_k}$ is the agent policy with parameters $\theta$ at step $k$. At the special case where there is no step $s$ that completes the chronic $c$ we define the default value $S2C(c) = |S_{TRAIN}|$, which is the worst possible value for this metric. As we can check in \eqref{eq:s2c}, the lower this value, the earlier the agent converges and reaches the desired stability using fewer trajectories and observations, which results in higher efficiency in this sense.

\begin{figure}[H]
    \centering
    \makebox[\textwidth][c]{%
      \includegraphics[width=\textwidth]{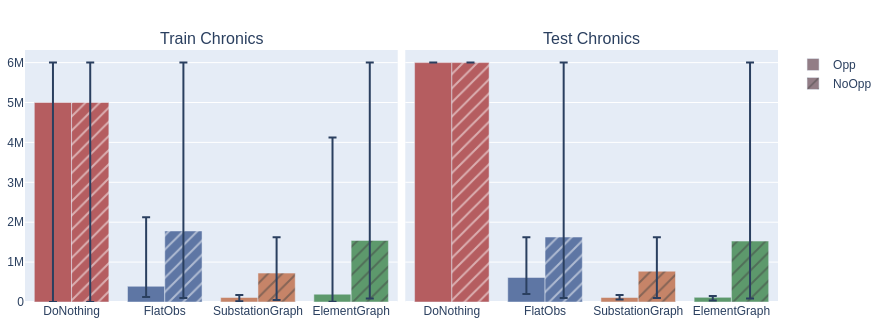}
    }
    \caption{Mean values of Steps to Complete metric of the proposed configurations for each agent trained with or without opponent (left and right bars respectively) and evaluated on training or test chronics (first and second plot respectively). The error bars indicate the maximum and minimum value obtained in 5 repetitions of the same configuration. DoNothing is included to the comparison to provide a baseline, helping to contextualize the performance of the other methods. It highlights the impact of the proposed configurations by illustrating how a non-trained agent performs in the same setup.}
    \label{fig:steps2complete}
\end{figure}

\begin{table}
    \tiny
    \centering
    \caption{Results of the S2C metric in million steps for each proposed configuration, training with and without opponent. These results are shown as the mean and standard deviation of 5 runs for each combination (DN:DoNothing, FO:FlatObs, SG:SubstationGraph, EG:ElementGraph). In bold is the winning combination for that particular chronic (Ch) and underlined is the second best.}
    \begin{tabular}{ll||c|ccc|ccc}
\toprule
 & Opp &  & \multicolumn{3}{c}{Trained without Opponent} & \multicolumn{3}{c}{Trained with Opponent} \\
 & Obs & DN & EG & FO & SG & EG & FO & SG \\
Ch & Set &  &  &  &  &  &  &  \\
\midrule
00 & Train & 6.0(±.0) & 1.6(±2.47) & 1.34(±2.61) & .78(±.7)  & \textbf{.12(±.03)} & .25(±.15) & \underline{.12(±.04)} \\
01 & Train & 6.0(±.0) & 1.6(±2.47) & 1.5(±2.54) & .78(±.7)  & \textbf{.11(±.04)} & .17(±.03) & \underline{.12(±.04)} \\
02 & Train & \textbf{.0(±.0)} & 1.6(±2.47) & 1.49(±2.55) & .52(±.66)  & \underline{.11(±.04)} & .55(±.88) & .12(±.04) \\
03 & Train & 6.0(±.0) & 1.45(±2.57) & 2.54(±2.24) & .76(±.72)  & \textbf{.1(±.04)} & .43(±.47) & \underline{.1(±.05)} \\
04 & Train & 6.0(±.0) & 1.6(±2.47) & 2.34(±2.43) & .76(±.71)  & \textbf{.11(±.04)} & \underline{.61(±.53)} & \textbf{.11(±.04)} \\
05 & Train & 6.0(±.0) & 1.6(±2.47) & 1.32(±2.62) & .78(±.7)  & \textbf{.11(±.04)} & .15(±.03) & \underline{.12(±.04)} \\
06 & Train & 6.0(±.0) & 1.6(±2.47) & 1.32(±2.61) & .78(±.7)  & \textbf{.11(±.04)} & .18(±.03) & \underline{.12(±.04)} \\
07 & Train & 6.0(±.0) & 1.45(±2.57) & 1.91(±2.62) & .76(±.72)  & \textbf{.1(±.04)} & .34(±.37) & \underline{.11(±.04)} \\
08 & Train & 6.0(±.0) & 1.45(±2.57) & 2.12(±2.48) & .76(±.72)  & \textbf{.1(±.04)} & .66(±.45) & \underline{.1(±.05)} \\
09 & Train & \textbf{.0(±.0)} & 1.45(±2.57) & 2.12(±2.48) & .5(±.67)  & \underline{.09(±.05)} & .47(±.46) & .1(±.05) \\
10 & Train & 6.0(±.0) & 1.6(±2.47) & 1.32(±2.61) & .78(±.7)  & \textbf{.11(±.04)} & .59(±.44) & \underline{.12(±.04)} \\
11 & Train & 6.0(±.0) & 1.45(±2.57) & 2.11(±2.49) & .76(±.72)  & .9(±1.8) & \underline{.4(±.41)} & \textbf{.1(±.06)} \\
12 & Train & 6.0(±.0) & 1.6(±2.47) & 1.5(±2.54) & .78(±.7)  & \textbf{.11(±.04)} & .2(±.05) & \underline{.12(±.04)} \\
13 & Train & \textbf{.0(±.0)} & 1.45(±2.57) & 2.54(±2.24) & .5(±.67)  & .89(±1.81) & .84(±.84) & \underline{.1(±.06)} \\
14 & Train & 6.0(±.0) & 1.6(±2.47) & 2.0(±2.47) & .78(±.7)  & \textbf{.11(±.04)} & .15(±.03) & \underline{.12(±.04)} \\
15 & Train & 6.0(±.0) & 1.6(±2.47) & 1.32(±2.61) & .78(±.7)  & \textbf{.12(±.03)} & .18(±.05) & \underline{.13(±.04)} \\
16 & Train & 6.0(±.0) & 1.6(±2.47) & 1.32(±2.61) & .78(±.7)  & \textbf{.11(±.04)} & .5(±.4) & \underline{.12(±.04)} \\
17 & Test & 6.0(±.0) & 1.45(±2.57) & 1.91(±2.63) & .77(±.7)  & \textbf{.11(±.04)} & \underline{.57(±.4)} & \textbf{.11(±.04)} \\
18 & Train & 6.0(±.0) & 1.45(±2.57) & 1.91(±2.63) & .76(±.72)  & \textbf{.1(±.04)} & .42(±.36) & \underline{.11(±.04)} \\
19 & Test & 6.0(±.0) & 1.6(±2.47) & 1.34(±2.6) & .78(±.7)  & \textbf{.12(±.03)} & .66(±.6) & \underline{.12(±.04)} \\

\bottomrule
\end{tabular}

    \label{tab:steps2complete}
\end{table}

In Figure~\ref{fig:steps2complete} and Table~\ref{tab:steps2complete} we observe that the stability metric $S2C$ has better values (both in training and test chronics) in agents training using an opponent and with observations that are graphs (ElementGraph and SubstationGraph). This result becomes more interesting when we compare it with the results shown from the training phase in Figure \ref{fig:train_rewards} since we can check that the increased difficulty of introducing an opponent in the training phase translates into greater ease of convergence and improved stability of the agent, while it might seem the opposite at first glance when seeing the results in Figure \ref{fig:train_rewards}.

Also, it can be seen that using an opponent during training and viewing the observation as a graph of elements is the best option except in cases where the agent is able to complete the chronic without doing anything (chronics 02, 09 and 13). It can also be seen how, when comparing under the same configuration, formalizing the observation as a dynamic graph of unique connections, or SubstationGraph, is also very useful for solving this problem. 

These results indicate that agents using a minimal amount of prior expert knowledge of the problem can perform well enough and that the proposed new form of action space is effective in solving a problem of this style even in unseen situations such as chronics 17 and 19, especially when combined with correct observation formalization as an element graph and using an opponent during training.

\subsection{L2RPN scores}

A common issue with agents trained through reinforcement learning is their excessive focus on optimizing their target reward, sometimes producing poor results in other aspects relevant to the problem at hand but not reflected in the reward function, a behavior sometimes known as "reward farming".  In order to provide another point of analysis based on an alternative objective function, we resort to a scoring metric employed in the aforementioned L2RPN competition. This metric takes into account both the costs related to the operation of the network and the cost associated with the network being down at a given point. 

This combined cost is calculated as 
\begin{equation} \label{eq:score}
    Cost_c(\pi) = \sum_{t = 0}^{t_{end}} CostOperation_t(c, \pi) +  \sum_{t = t_{end}}^{|c|} CostBlackout_t(c),
\end{equation}
where $\pi$ is the agent's policy to evaluate, $c$ and $|c|$ are the chronic and it's size in steps respectively, and $t_{end}$ is the step that the agent falls into a blackout, if that happens. 

As this particular environment has neither re-dispatching, storages nor curtailments we assume those costs as zero, so we have that 
\begin{equation*}
    CostOperation_t(c, \pi) = EnergyLoss_t(c, \pi) \cdot EnergyPrice_t(c)
\end{equation*}
and 
\begin{equation*}
    CostBlackout_t(c) = \beta \cdot \sum_{l \in L}load_l(t) \cdot EnergyPrice_t(c),
\end{equation*}
where $\beta > 1$ is a penalty for blackout states and $L$ is the load set.

Using this common cost, the cost of the ``Do Nothing'' agent can be calculated in each of the chronics and used as a baseline. Following this idea, the L2RPN competition defines a score in the following way: an agent attaining the same cost as the ``Do Nothing'' agent is awarded  $score = 0.0$, while a saving of $80\%$ over such baseline costs is awarded $score = 100$ . Other levels of cost savings are awarded a score following a linear transformation between these two points, e.g. a saving of a $40\%$ would attain $score = 50$.

\begin{figure}[H]
    \centering
    \includegraphics[width=1\textwidth]{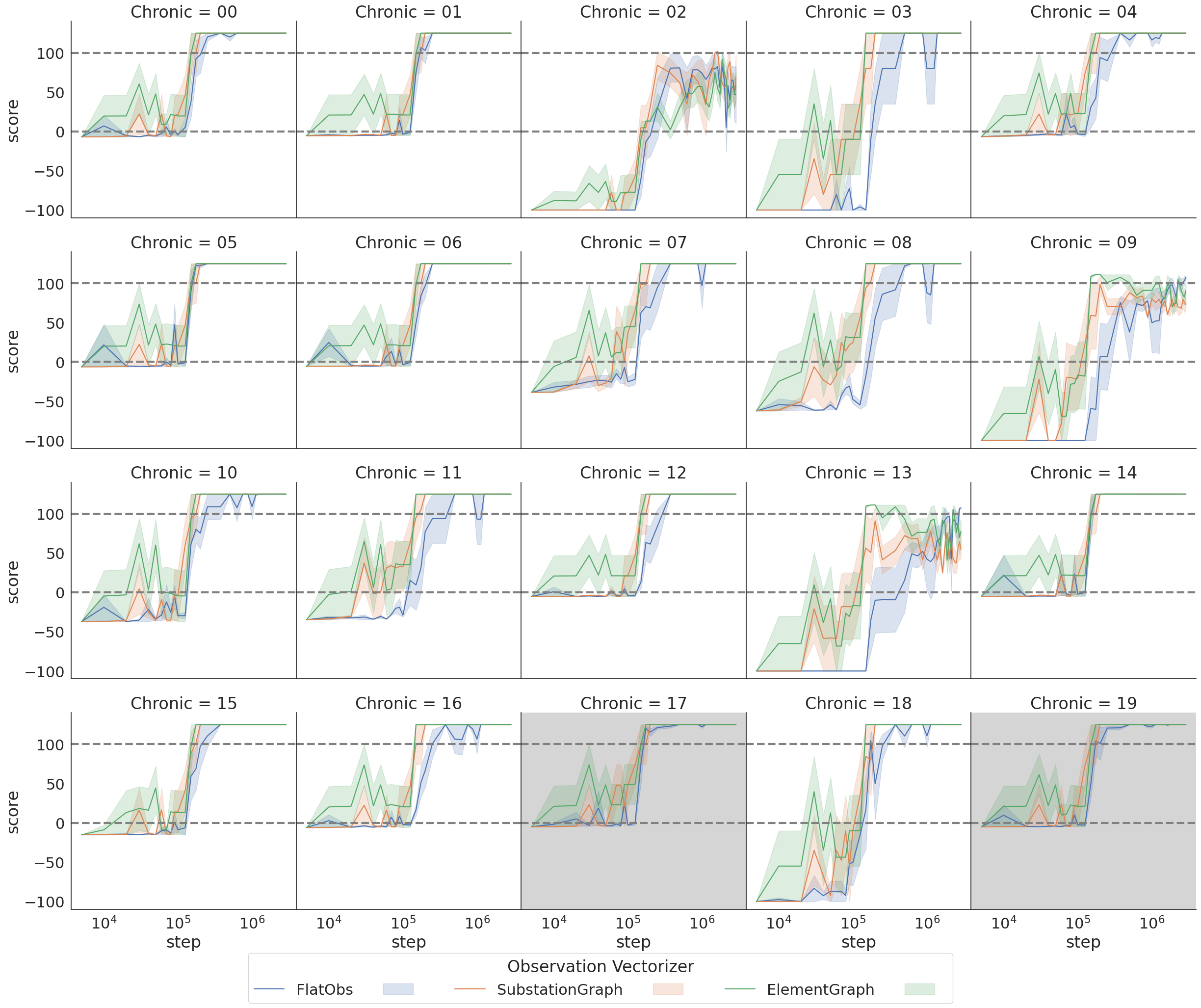}
    \caption{Agents' scores computed following the methodology proposed in the L2RPN competition \ref{eq:score}. In this context, scores of 0.0 and 100 correspond to the cost of the ``Do Nothing'' agent and 20\%  savings over that cost, respectively. Grayed-out chronics correspond to chronics not used during training. The figure highlights the superior convergence of element-graph-based agents, despite the competitiveness of other approaches. These results confirm that the strategies proposed in this paper are effective for solving this problem.}
    \label{fig:scores}
\end{figure}

In this way, we can obtain the graphs in Figure \ref{fig:scores}, in which we can see that the agents trained with the proposed strategy are even able to improve the $80\%$ cost savings of reaching $score=100$ in many chronics, including cases never seen before, such as chronics 17 and 19.
\section{Conclusions and Future Work} \label{sec:concl}
In this paper, we presented a novel method to address a well-known problem, revisiting it through a perspective more aligned with a classical model-free reinforcement learning framework. In order to achieve this goal, a masked topological action space has been designed to allow the agent full control over the network without the need to make use of expert knowledge 

Different formalizations of observations have also been explored, testing how a power grid can be viewed as a graph of elements when making decisions about it in a more robust way that converges faster to stable regimes. 

The advances discussed above have allowed to obtain agents that are able to operate a network in all the proposed cases, also being able to generalize to situations not seen before in their training (chronics 17 and 19) while minimizing the cost of power losses along the network when compared to ``Do Nothing'' baseline. 

These results open up a wide range of possibilities for future work. First, it is necessary to extend the conclusions obtained in this network topology to larger networks, either of 14 or 118  substations (already available as scenarios in the Grid2Op framework). On the other hand, it can also be investigated how to add different terms that meet the conditions specified for the reward function so that the agent can be more efficient in other aspects as well (priority use of renewable energies, efficient use of batteries, minimization of electricity costs to consumers, etc.). Finally, it is intended to continue developing the formalization of the ElementGraph formalization proposed in this paper, which is flexible enough to make a policy that does not need to vectorize the observation and can make a direct mapping of the elements at the input with their corresponding actions at the output. This would allow to obtain an agent that is able to manage several network topologies simultaneously, and thus could be the way to build a possible foundational model for power grid management.

\section{Acknowledgments}
The authors acknowledge financial support from project PID2022-139856NB-I00 funded by MCIN/ AEI / 10.13039/501100011033 / FEDER UE, from Comunidad de Madrid e IDEA-CM (TEC-2024/COM-89), the Instituto de Ingeniería del Conocimiento (IIC) and the project IA4TES - Inteligencia Artificial para la Transición Energética Sostenible funded by Ministry of Economic Affairs and Digital Transformation.

\section{Credit assignment}
The code of this article and its writing has been done entirely by Eloy Anguiano Batanero under the supervision of Ángela Fernández Pascual and Álvaro Barbero Jiménez.

\bibliography{ms}{}
\bibliographystyle{plainnat}

\newpage

\begin{appendices} \label{appendixes}

\section{Environments configuration and model hyperparameters.}
The Grid2Op framework offers the ability to modify the rules of all the environments it contains in order to train different agents for different situations that may occur. In our case, we have made modifications to these rules to enable total freedom of the agent when making decisions, and to establish a specific difficulty based on the existing literature. 

In the first place, some modifications that seek to provide the agent with greater freedom in the environment have been made in order to take advantage of the action space proposed in this work. These are the parameters that control whether the environment should enable actions that change substations or lines simultaneously. These parameters have been set to a sufficiently high number (9999 in both cases), so that they never suppose a limitation to the agents proposed here.

On the other hand, there are other parameters that manage the difficulty of the environment itself. We have based our settings on the configuration used in \citet{MARL} which sets the following values in the code found in \citet{MARLGit}: 

\begin{table}[H]
\tiny
\begin{tabular}{l|l|p{6cm}}
\toprule
Parameter                       & Value & Description \\
\midrule
MAX\_SUB\_CHANGED & 9999     & Maximum number of substations that can be reconfigured between two consecutive timesteps by an agent.\\
\midrule
MAX\_LINE\_STATUS\_CHANGED & 9999     & Maximum number of powerlines statuses that can be changed between two consecutive timesteps by an agent.\\
\midrule
NB\_TIMESTEP\_OVERFLOW\_ALLOWED & 3     & Number of timesteps for which a soft overflow is allowed. This means that a powerline will be disconnected after 3 time steps above its thermal limit.\\
\midrule
NB\_TIMESTEP\_RECONNECTION      & 12    & Number of timesteps a powerline disconnected for security motives. \\
\midrule
NB\_TIMESTEP\_COOLDOWN\_LINE    & 3     & When an agent acts on a powerline by changing its status (connected / disconnected) this number indicates how many timesteps the agent has to wait before being able to modify this status again. \\
\midrule
NB\_TIMESTEP\_COOLDOWN\_SUB     & 3     & When an agent changes the topology of a substations, this number indicates how many timesteps the agent has to wait before being able to modify the topology on this same substation. \\
\midrule
HARD\_OVERFLOW\_THRESHOLD       & 200   & If the powerflow on a line is above HARD\_OVERFLOW\_THRESHOLD * thermal\_limit then it is automatically disconnected, regardless of the number of timesteps it is on overflow.\\
\bottomrule

\end{tabular}
\end{table}

Finally, we have the hyperparameters relative to each of the models. First, after each observation vectorization, we have two different flows for the policy and the value function respectively. In both cases, we have a fully connected three-layer network of 128 neurons. Subsequently, the FlatObs adds a single layer of 128 neurons to the vector representing the observation, while in the case of the graph observations (SubstationGraph and ElementGraph) the GraphTransformer configuration is quite simple and a hidden size of 128 is used with a single encoder layer and a single attention head. It is worth noting that in the case of ElementGraph enables to add dense layers of specialization at the input and output of each of the elements, so we would be talking about two extra layers of 128 neurons also for each of the types of elements of the graph. Also, every agent has been trained with an entropy bonus of $0.01$.

\section{Additional results} \label{appendix:results}
Following the results presented in the Table \ref{tab:steps2complete} we can easily conclude that the training configuration of the agents that yields the best results are those that are trained with an opponent. However, it is interesting to see the improvements acquired throughout their training period in deterministic mode (with exploration disabled) for each of the complete chronicles (2016 steps), either with or without the influence of an opponent.

\begin{figure}[H]
    \centering
        \includegraphics[width=\textwidth]{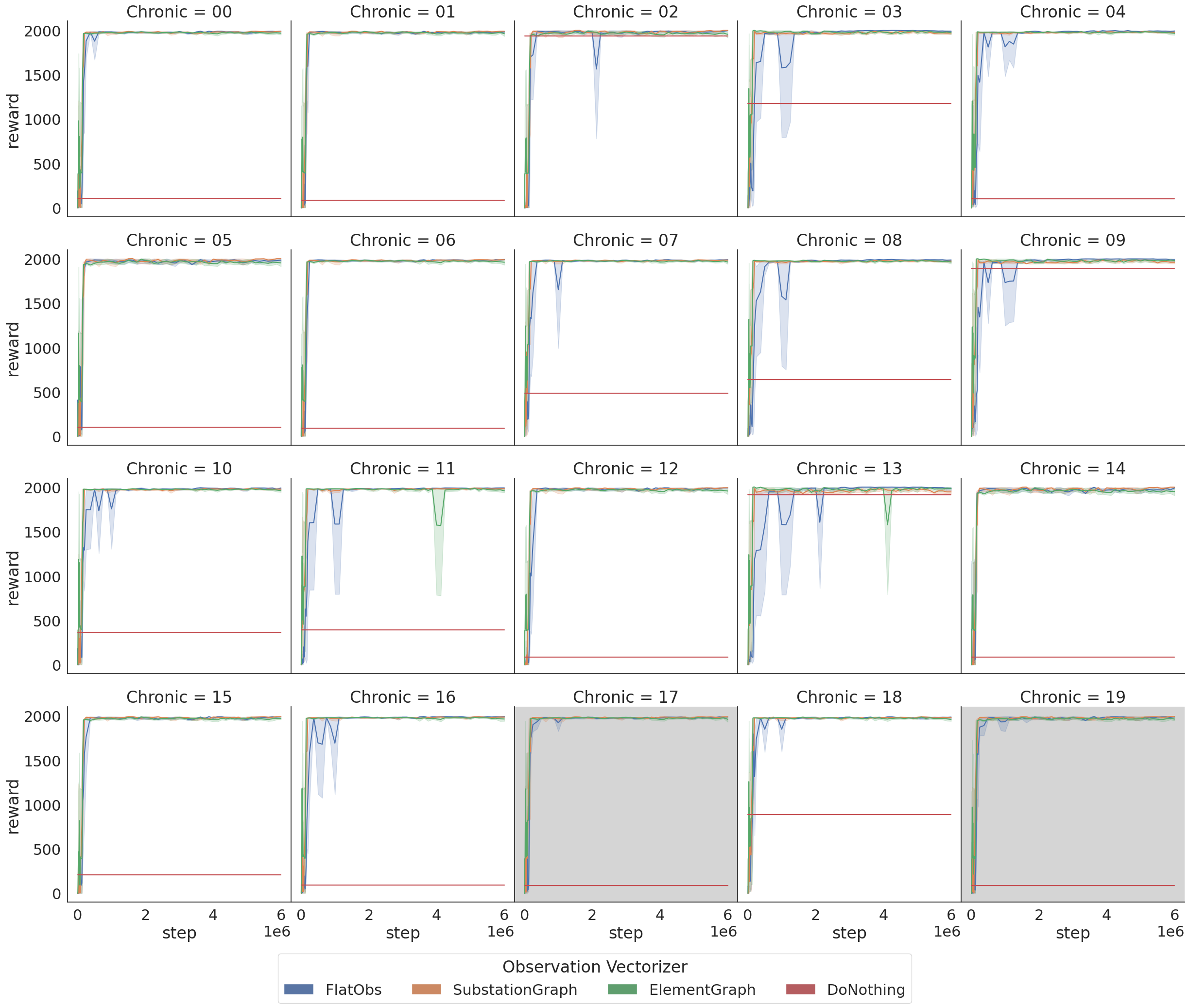}
    \caption{Rewards in each of the chronics without opponent of the different agent configurations averaged over 5 runs. Chronics 17 and 19 are shaded to indicate that they were not part of the training set.}
    \label{fig:test_rew_noop}
\end{figure}

\begin{figure}[H]
    \centering
        \includegraphics[width=\textwidth]{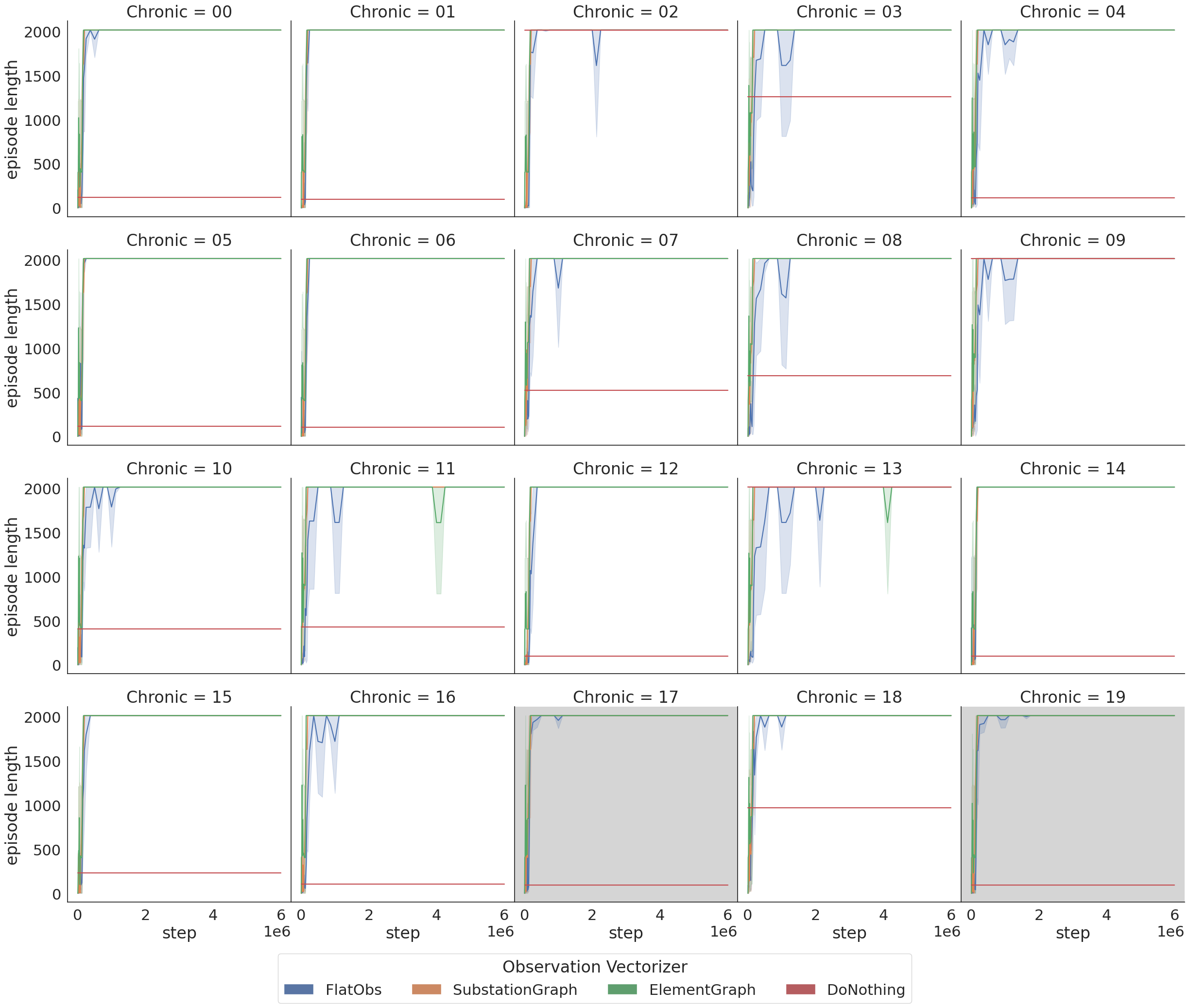}
    \caption{Episode lengths in each of the chronics without opponent of the different agent configurations averaged over 5 runs. Chronics 17 and 19 are shaded to indicate that they were not part of the training set.}
    \label{fig:test_len_noop}
\end{figure}

\begin{figure}[H]
    \centering
        \includegraphics[width=\textwidth]{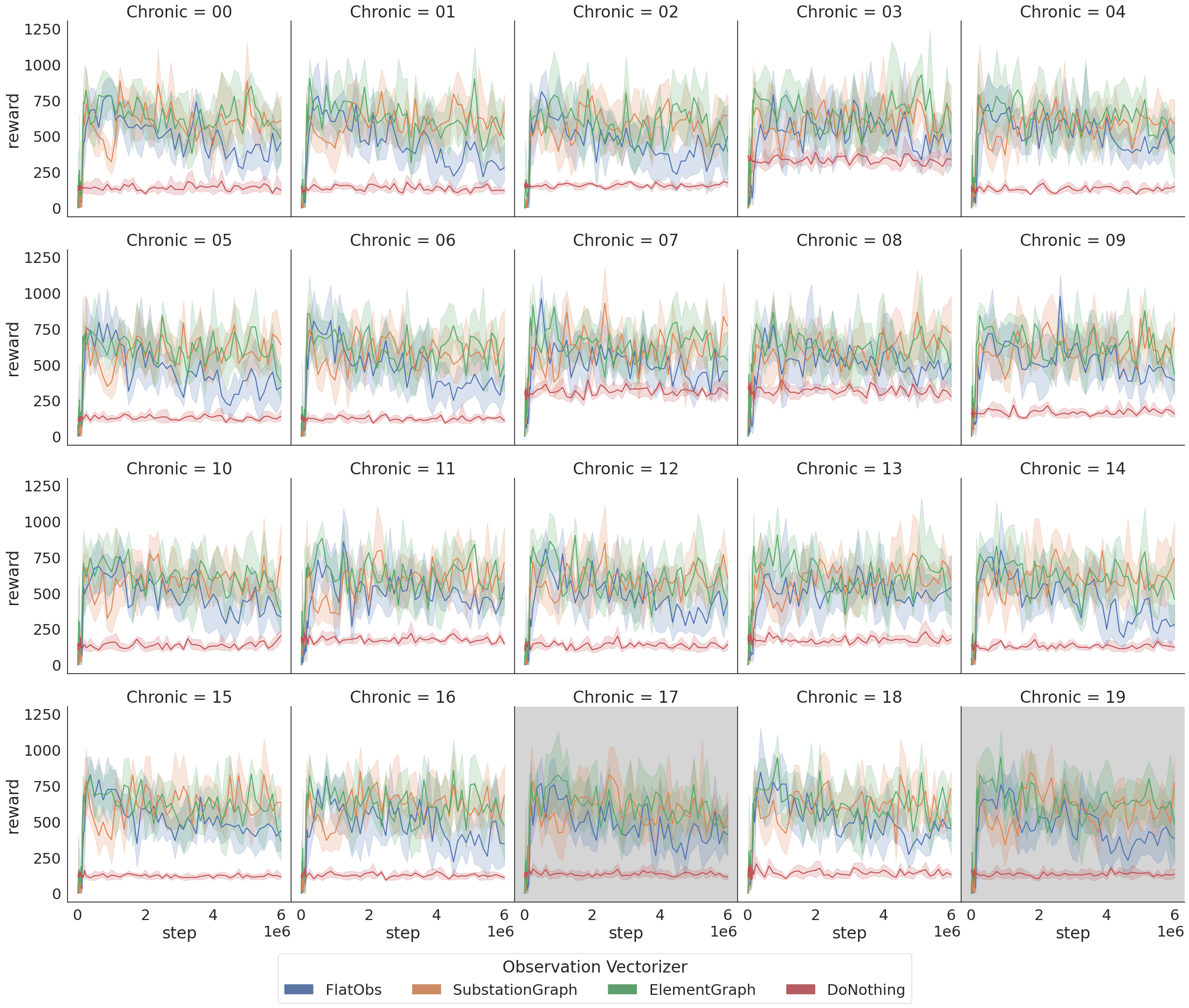}
    \caption{Rewards in each of the chronics with opponent of the different agent configurations averaged over 5 runs. Chronics 17 and 19 are shaded to indicate that they were not part of the training set.}
    \label{fig:test_rew_opp}
\end{figure}

\begin{figure}[H]
    \centering
        \includegraphics[width=\textwidth]{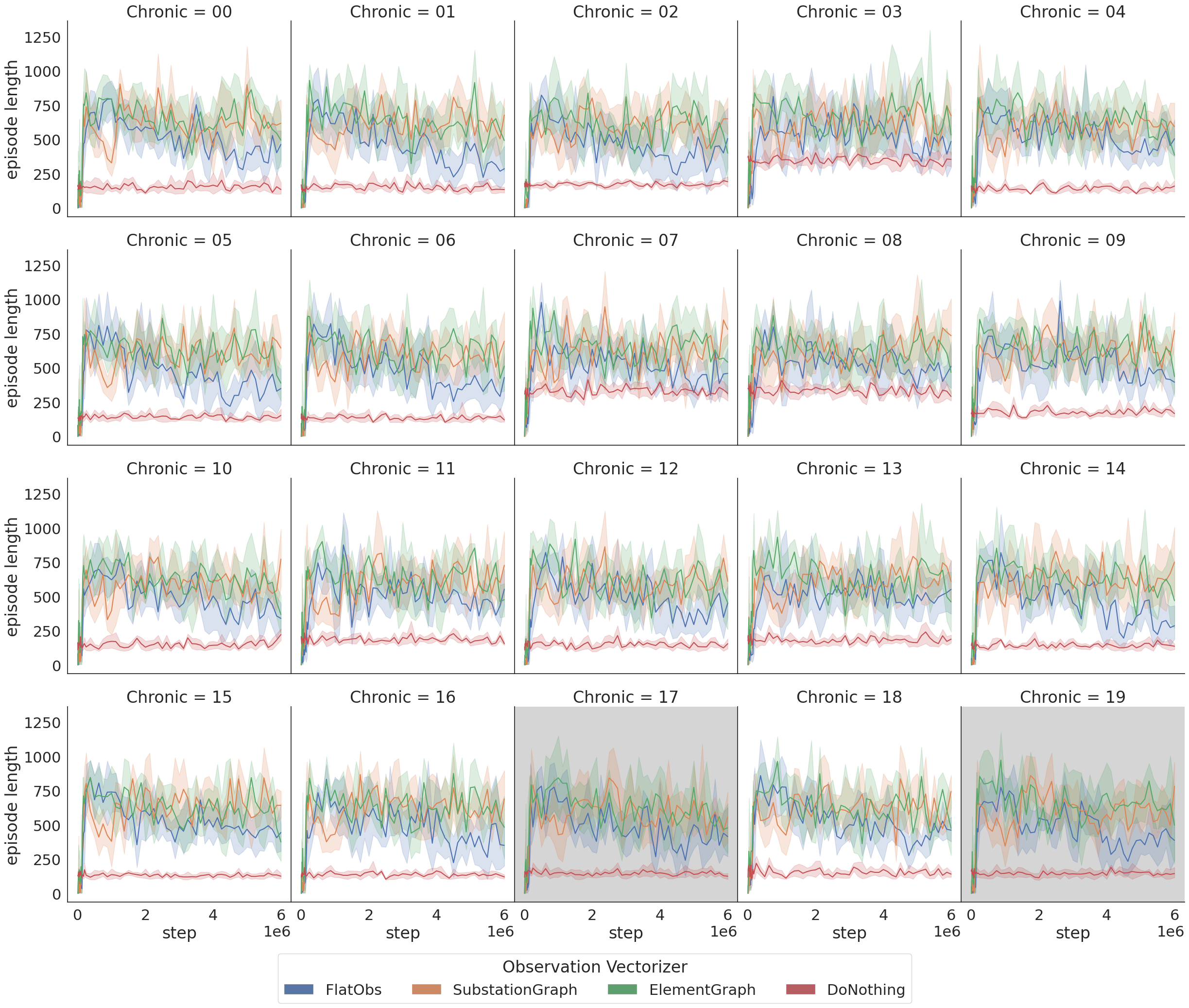}
    \caption{Episode lengths in each of the chronics with opponent of the different agent configurations averaged over 5 runs. Chronics 17 and 19 are shaded to indicate that they were not part of the training set.}
    \label{fig:test_len_opp}
\end{figure}

Thus, the results of the periodic tests during the training process of the agents that have been trained under the best configuration found are presented at Figures \ref{fig:test_rew_noop}, \ref{fig:test_len_noop}, \ref{fig:test_rew_opp} and \ref{fig:test_len_opp}.

\end{appendices}

\end{document}